\documentclass[journal]{vgtc} 
\usepackage{amsmath}
\usepackage{booktabs}
\usepackage{pifont}
\newcommand{\cmark}{\ding{51}}
\newcommand{\xmark}{\ding{55}}
\usepackage{makecell}
\usepackage{booktabs}
\usepackage[table]{xcolor}
\usepackage{booktabs}
\usepackage{multirow}
\usepackage{array}
\onlineid{0}

\vgtccategory{Research}

\title{ChartFI: Benchmarking Faithfulness and Insightfulness of Chart Descriptions from Multimodal Large Language Models}

\author{%
  \authororcid{Fen Wang}{0009-0002-1039-2813},
  \authororcid{Zekai Shao}{0000-0003-2014-5293},
  Qiman Kang,
  Chunran Hu,
  Zhixuan Zhang,
  Lexu Xie,
  \authororcid{Chao Liu}{0000-0002-9748-3162}, and
\authororcid{Siming Chen}{0000-0002-2690-3588}
}

\authorfooter{
  \item
  	Fen Wang, Zekai Shao, Qiman Kang, Chunran Hu, Zhixuan Zhang, Siming Chen are with School of Data Science, Fudan University. Siming Chen is the corresponding author. E-mail: simingchen@fudan.edu.cn.
    \item Fen Wang and Zekai Shao contributed equally as first authors.
  \item Lexu Xie is with Zhengzhou Zhongke Institute of Integrated Circuit and System Application.
  \item
  	Chao Liu is with School of Computer Science, Fudan University and Zhengzhou Zhongke Institute of Integrated Circuit and System Application.

}

\abstract{
Chart descriptions are essential for accessibility, cross-modal retrieval, and assisting readers in extracting insights from complex visualizations. As multimodal large language models (MLLMs) are increasingly adopted for automated chart description generation, a critical question arises: how faithfully and insightfully do these models actually describe charts?  Current benchmarks fall short on two fronts: existing datasets consist of simple, homogeneous charts paired with shallow, fact-enumerating descriptions; and prevailing metrics fail to capture the multi-faceted nature of description quality. To address these gaps, we present the Chart Faithfulness and Insightfulness Benchmark (ChartFI-Bench). We first summarize four dimensions that characterize high-quality chart descriptions: factual accuracy, salient feature emphasis, domain-informed guidance, and chart–text complementarity. Guided by these dimensions, we construct a high-quality benchmark comprising 896 chart--description pairs, which feature visually complex charts and semantically rich descriptions. Furthermore, we design four aligned evaluation metrics---Faithfulness, Coverage, Informativeness, and Acuity---to systematically assess the quality of descriptions across these dimensions. Experiments conducted on mainstream MLLMs demonstrate the effectiveness of the proposed framework and reveal common weaknesses among existing models.

}

\keywords{Chart Description, Evaluation Metric, Benchmark, Multimodal large language models}
\graphicspath{{figs/}{figures/}{pictures/}{images/}{./}} % where to search for the images

\usepackage{tabu}                      % only used for the table example
\usepackage{booktabs}                  % only used for the table example
\usepackage{lipsum}                    % used to generate placeholder text
\usepackage{mwe}                       % used to generate placeholder figures
\usepackage{ccicons}                   % package to be able to use icons from creative commons

\usepackage{mathptmx}                  % use matching math font

\begin{document}

%%%%%%%%%%%%%%%%%%%%%%%%%%%%%%%%%%%%%%%%%%%%%%%%%%%%%%%%%%%%%%%%
%%%%%%%%%%%%%%%%%%%%%% START OF THE PAPER %%%%%%%%%%%%%%%%%%%%%%
%%%%%%%%%%%%%%%%%%%%%%%%%%%%%%%%%%%%%%%%%%%%%%%%%%%%%%%%%%%%%%%%

%% The ``\maketitle'' command must be the first command after the
%\\begin{abstract}
% ``\begin{document}'' command. It prepares and prints the title block.
%% the only exception to this rule is the \firstsection command
\firstsection{Introduction}
  
\maketitle

High-quality chart descriptions reduce cognitive load\cite{latif2021kori}, facilitating rapid comprehension of the core content and assisting readers in discovering deeper insights underlying the data~\cite{vistext,chart-to-text,wang2025chartinsighter}. However, the automatic generation of such descriptions necessitates the integration of multiple complex tasks, including the perception of visual features, the mapping between graphical elements and numerical values, and the application of domain knowledge for reasoning. Recently, Multimodal Large Language Models (MLLMs) have demonstrated strong capabilities in visual understanding, thereby advancing research in automated generation of chart descriptions. For instance, VisText~\cite{vistext} and ChartCap~\cite{lim2025chartcap} train chart-specific models to generate chart descriptions, whereas ChartInsighter~\cite{wang2025chartinsighter} and Vistoryteller~\cite{vistoryteller} leverage Large Language Models (LLMs) to invoke external tools for mitigating hallucinations in the generated text.

% Although prior studies have explored the generation of high-quality chart descriptions from diverse methodological perspectives, current datasets mostly comprise simple, stylistically homogeneous charts paired with descriptions that enumerate surface-level facts without conveying deeper analytical insights. However, evidence suggests that readers prefer descriptions that provide deep analytical insights~\cite{fourlevel,vistext}. This gap between the characteristics of datasets and the expectations of readers makes it difficult to assess whether current methods can produce descriptions that meet real-world information needs. 

Prior studies have explored the generation of high-quality chart descriptions from diverse methodological perspectives. \textcolor{black}{It remains unclear whether current models can reliably generate descriptions that are both faithful to the underlying data and analytically insightful. This uncertainty persists because existing evaluation frameworks suffer from two limitations.} On the data side, current benchmarks mostly comprise simple, stylistically homogeneous charts paired with descriptions that enumerate surface-level facts without conveying deeper analytical insights. On the metric side, traditional natural language generation metrics such as BLEU~\cite{papineni2002bleu} and METEOR~\cite{banerjee2005meteor} \textcolor{black}{primarily quantify lexical overlap between generated and reference descriptions}, rendering them unreliable for assessing factual correctness or analytical depth. \textcolor{black}{ For example, semantically opposite statements such as ``A outperforms B'' and ``B outperforms A'' may still receive high scores because of their substantial lexical overlap.}

% share most of their surface tokens yet carry opposite meanings, yet these measures assign inflated scores to semantically contradictory statements.

\textcolor{black}{Although recent frameworks have begun to address the limitations of traditional reference-based metrics, their applicability to the evaluation of chart descriptions remains limited. Metrics developed for general image captioning~\cite{CAPTURE,liu2025capability} are not well suited to evaluating chart descriptions because they partition descriptions into relatively independent words or phrases and treat them as separate evaluation units. However, chart descriptions often exhibit strong semantic dependencies, and such decomposition can disrupt the overall semantic structure, preventing the resulting scores from accurately reflecting the quality of the generated descriptions. Chart-specific metrics~\cite{lim2025chartcap,huang2024lvlms} focus primarily on factual faithfulness, leaving insightfulness largely unaddressed.} More recent approaches that employ LLMs as evaluators~\cite{xia2025chartx} have not been rigorously validated for reliability in the chart domain, leaving their trustworthiness uncertain.

\begin{table*}[t]
\centering
\caption{\textcolor{black}{Comparison of ChartFI-Bench and Existing Datasets. Semantic Coverage refers to the four-level semantic content for chart descriptions\cite{fourlevel}.}}
\fontsize{6.75pt}{8pt}\selectfont 
\label{tab:dataset_comparison}
% \footnotesize
% \tiny
\renewcommand{\arraystretch}{0.95}
\setlength{\tabcolsep}{4pt}
\resizebox{\textwidth}{!}{%
\begin{tabular}{lccccccccc}
\toprule
\multirow{2}{*}[-0.3ex]{\textbf{Dataset}}
& \multirow{2}{*}[-0.3ex]{\textbf{Size}}
& \multirow{2}{*}[-0.3ex]{\textbf{Word Count}}
& \multirow{2}{*}[-0.3ex]{\begin{tabular}[c]{@{}c@{}}\textbf{Chart}\\\textbf{Types}\end{tabular}}
& \multirow{2}{*}[-0.3ex]{\begin{tabular}[c]{@{}c@{}}\textbf{Semantic}\\\textbf{Coverage}\end{tabular}}
& \multirow{2}{*}[-0.3ex]{\begin{tabular}[c]{@{}c@{}}\textbf{Multi-}\\\textbf{Chart}\end{tabular}}
& \multicolumn{4}{c}{\textbf{Evaluation Dimensions}} \\ 
\cmidrule(lr){7-10}
& & & & & &\textbf{ Faithfulness} & \textbf{Coverage} & \textbf{Informativeness }& \textbf{Acuity }\\
\midrule
Chart-to-Text\cite{chart2text_image} 
& 44K & 120 & 6 & L1--L3 & \xmark 
& \xmark & \cmark & \xmark & \xmark \\

ChartSumm\cite{rahman2023chartsumm} 
& 84K & 45 & 3 & L1--L3 & \xmark 
& \xmark & \cmark & \xmark & \xmark \\

VisText\cite{vistext} 
& 12K & 77 & 3 & L1--L3 & \xmark 
& \cmark & \cmark & \xmark & \xmark \\

ChartLlama\cite{han2023chartllama} 
& 11K & - & 10 & L1--L3 & \xmark 
& \xmark & \cmark & \xmark & \xmark \\

VL2NL\cite{ko2024natural} 
& 2K & 161 & 10 & L1--L2 & \xmark 
& \xmark & \cmark & \cmark & \xmark \\

ChartInsighter\cite{wang2025chartinsighter} 
& 75 & 157 & 1 & L1--L3 & \xmark 
& \xmark & \cmark & \cmark & \xmark \\

ChartCap\cite{lim2025chartcap} 
& \textbf{565K} & 223 & 9 & L1--L3 & \cmark 
& \cmark & \cmark & \xmark & \xmark \\

ChartFI (Ours)
& 896 & \textbf{241} & \textbf{14} & \textbf{L1--L4} & \cmark 
& \cmark & \cmark & \cmark & \cmark \\
\bottomrule
\end{tabular}
}
\vspace{-10pt} %
\end{table*}

To address these gaps, we propose the Chart Faithfulness and Insightfulness Benchmark (ChartFI-Bench), a comprehensive benchmark for evaluating MLLM-generated chart descriptions \textcolor{black}{along these two dimensions.} To construct the benchmark, we first summarize four core characteristics of high-quality chart descriptions: factual accuracy, salient feature emphasis, domain-informed guidance and chart--text complementarity. \textcolor{black}{These characteristics guide both benchmark construction and evaluation methodology design. For benchmark construction,} we collect charts from arXiv and apply a systematic filtering pipeline followed by manual verification, ultimately obtaining 896 chart-description pairs with visually complex charts and semantically rich descriptions. 

\textcolor{black}{For evaluation methodology design}, we construct a four-dimensional evaluation framework to systematically quantify the faithfulness and insightfulness of chart descriptions. \textcolor{black}{Specifically, we first decompose each description into fine-grained insights \textcolor{black}{for more precise evaluation}. For Faithfulness evaluation, we compare four verification strategies and adopt a direct MLLM-as-a-Judge approach, which performs best among the evaluated strategies.} \textcolor{black}{For insightfulness evaluation, we further assess chart descriptions along three dimensions: Coverage, Informativeness, and Acuity.} For Coverage, we match reference and generated \textcolor{black}{chart insights} through insight-type-specific scoring rules to enable precise alignment. We further introduce an Informativeness metric with context-adaptive weighting that adjusts semantic-level importance based on chart complexity, and an Acuity metric \textcolor{black}{that assesses the model's ability to apply domain knowledge across five sub-dimensions.} Subsequently, we utilize the constructed benchmark and the proposed evaluation framework to conduct a comprehensive evaluation of state-of-the-art MLLMs. \textcolor{black}{The evaluation results reveal common weaknesses among current models and suggest directions for future improvement.} \textcolor{black}{We also quantify the consistency between the proposed metrics and human judgments to assess the effectiveness of the automatic evaluation.} The benchmark is available at 
\href{https://github.com/wangfen01/ChartFI}{https://github.com/wangfen01/ChartFI}. Our main contributions are as follows:
\begin{itemize}[itemsep=0.5pt, topsep=0.2pt, parsep=0pt, partopsep=0pt]
    \item \textcolor{black}{We summarize the characteristics of high-quality chart descriptions in terms of faithfulness and insightfulness, and identify four dimensions for benchmark construction and evaluation design.}
    \item \textcolor{black}{We construct a high-quality benchmark of 896 chart-description pairs and propose an evaluation framework for assessing the faithfulness and insightfulness of chart descriptions.}
    \item \textcolor{black}{We conduct comprehensive evaluations of state-of-the-art MLLMs and reveal common weaknesses among existing models.}
\end{itemize}

\section{Related Work}

This section reviews prior research on chart description, focusing on large models, benchmark datasets, and evaluation metrics.
\subsection{Large Models for Chart Description}

 \textcolor{black}{Lundgard and Satyanarayan~\cite{fourlevel} proposed a four-level semantic framework for chart descriptions: Level 1 (L1) describes chart construction (e.g., \textit{``The graph plots the number of people on the x-axis against movie genres on the y-axis''}); Level 2{ (L2) provides statistical summaries (e.g., \textit``Sales peaked at 500 units''}); Level 3{ (L3) highlights perceptual and cognitive observations (e.g., \textit``Carbon emission levels exhibit high volatility across the decade''}); and Level 4 {(L4) presents contextual and domain-specific insights (e.g., \textit``The sharp decline in global trade volume is attributed to the onset of a specific economic recession''}). Recently, an increasing number of studies have applied large models to generate chart descriptions.} Early research on chart description primarily relied on template-based generation methods~\cite{hsu2021scicap,chen2019figure,liu2024chartthinker}, which severely limit the diversity and flexibility of generated content. Subsequent work shifted toward data-driven generation paradigms. Chart-to-Text~\cite{chart-to-text} and DataTales~\cite{sultanum2023datatales} converted the underlying structured data of charts into textual descriptions; however, this approach depends solely on data table inputs and inevitably discards critical visual information such as color and layout. While LineCap~\cite{mahinpei2022linecap}, Chart2Text~\cite{chart2text_image} and MMCA\cite{liu2024mmc} incorporated visual features, their descriptions remain limited, often failing to capture the deeper trends and insights embedded in charts. To enhance the semantic richness of generated descriptions, VisText\cite{vistext} pursued this goal but is restricted to simple, single-dimensional chart types, making it difficult to handle complex charts. 

% VL2NL~\cite{ko2024natural} guided LLMs to focus on statistical features and utilized external tools for data analysis, partially alleviating numerical hallucinations. 

To address hallucinations in chart description, recent work has introduced external reasoning mechanisms to improve factual accuracy. \textcolor{black}{ChartInsighter~\cite{wang2025chartinsighter} incorporates external modules to support reasoning, enabling deeper analysis of trends while reducing hallucinations. Vistoryteller~\cite{vistoryteller} simulates a human-like authoring process to create cohesive data stories aligned with user-specified themes.} In summary, current methods have evolved along three dimensions: diversity, semantic depth, and factual accuracy. However, \textcolor{black}{existing benchmarks fall short in evaluating them under complex scenarios. To address this gap, we construct a comprehensive benchmark to better evaluate these methods on faithfulness and insightfulness.}
% These approaches demonstrated stronger generation capabilities but simultaneously faced challenges including massive parameter counts and prohibitive computational costs, while their generalization to complex charts remained insufficient.

\vspace{-4pt}
\subsection{Dataset for Chart Description}
\vspace{-2pt}
Early chart description datasets were constructed through rule-based or template-driven generation methods~\cite{han2023chartllama}. However, such synthetic data exhibits a considerable gap from real-world charts in both visual style and structural complexity. To address this limitation, several studies have collected real chart descriptions written by human experts. For instance, Chart-to-Text\cite{chart2text_image} aggregated authentic chart resources from websites such as Statista\cite{statista} and Pew\cite{pewresearch}. VisText\cite{vistext} further enriched the semantic coverage of descriptions by incorporating multi-level features that span statistical, perceptual, and cognitive dimensions\cite{fourlevel}. ChartInsighter\cite{wang2025chartinsighter} constructed chart descriptions for time-series data and annotated hallucination types of descriptions generated by different models at the sentence level, thereby facilitating the evaluation of hallucination in chart description generation. \textcolor{black}{Nevertheless, the charts in these datasets are mostly simple and homogeneous. SciCap~\cite{hsu2021scicap} and ArxivCap~\cite{Arxivcap} constructed large-scale, domain-specific chart datasets from scientific literature, yet their textual content, which was primarily extracted from figure captions, lacks a comprehensive semantic interpretation of the charts.}

With the advancement of LLMs, \textcolor{black}{a growing body of research has leveraged these models to automate the generation of charts and descriptions. For instance, ChartAssistant~\cite{meng2024chartassistant} and ChartLlama~\cite{han2023chartllama} employed LLMs to synthesize diverse datasets and generate the corresponding code for chart rendering. ChartCap~\cite{lim2025chartcap} and ChartDiff\cite{ye2026chartdiff} utilized carefully designed instructions to guide LLMs in generating descriptions of charts. However, existing approaches primarily enumerate surface-level statistical facts and lack deep, domain-grounded insights; furthermore, the generated text frequently exhibits hallucination. To address these limitations, we construct a high-quality benchmark for chart descriptions that integrates the knowledge of domain experts with rich insights. We ensure the quality of the benchmark through manual inspection. A systematic comparison is presented in Tab.~\ref{tab:dataset_comparison}.}

% \begin{table*}[t]
% \centering
% \caption{Comparison of ChartFI-Bench and existing benchmarks.}
% \label{tab:dataset_comparison}
% \footnotesize
% \renewcommand{\arraystretch}{1.15}
% \setlength{\tabcolsep}{4pt}
% \resizebox{\textwidth}{!}{%
% \begin{tabular}{lccccc>{\centering\arraybackslash}m{4.8cm}}
% \toprule
% \makecell[c]{Dataset}
% & \makecell[c]{Size}
% & \makecell[c]{Avg. Description\\Word Count}
% & \makecell[c]{Chart\\Types}
% & \makecell[c]{Semantic\\Coverage}
% & \makecell[c]{Multi-\\Chart}
% & \makecell[c]{Evaluation\\Metric} \\
% \midrule
% Chart-to-Text\cite{chart2text} 
% & 44K & 120 & 6 & L1--L3 & \xmark 
% & \makecell[c]{BLEU \& CS \& BLEURT \& CIDEr \& PPL} \\

% ChartSumm\cite{rahman2023chartsumm} 
% & 84K & 45 & 3 & -- & \xmark 
% & \makecell[c]{BLEU \& CS \& BLEURT \& CIDEr \& PPL} \\

% VisText\cite{vistext} 
% & 12K & 77 & 3 & L1--L3 & \xmark 
% & \makecell[c]{BLEU, PPL, RG,\\ROUGE, WMD, TER} \\

% VL2NL\cite{ko2024natural} 
% & 2K & -- & 10 & L1--L2 & \xmark 
% & \makecell[c]{Within-Distribution,\\Cross-Distribution} \\

% ChartInsighter\cite{wang2025chartinsighter} 
% & 75 & 157 & 1 & L1--L3 & \xmark 
% & \makecell[c]{Within-Distribution,\\Semantic Richness,\\Hallucination Rate} \\

% ChartCap\cite{lim2025chartcap} 
% & \textbf{565K} & 223 & 9 & L1--L3 & \cmark 
% & \makecell[c]{BLEU, ROUGE, METEOR,\\BERTScore, VCS, OCRScore} \\

% Ours 
% & 897 & \textbf{241} & \textbf{15} & \textbf{L1--L4} & \cmark 
% & BLEU \\
% \bottomrule
% \end{tabular}
% }
% \end{table*}

\subsection{Evaluation for Chart Description}

 \textcolor{black}{Classical metrics like BLEU~\cite{papineni2002bleu}, CIDEr~\cite{vedantam2015cider}, and BERTScore~\cite{zhang2019bertscore} rely predominantly on superficial lexical or embedding alignments. By inappropriately rewarding deceptive word overlap and penalizing valid stylistic variations, they struggle to reflect a model's true comprehension capabilities. CLIPScore~\cite{hessel2021clipscore} captures only high-level semantic alignment, rendering it ill-suited for lengthy, detail-rich chart narratives. Furthermore, metrics designed for general image captioning, such as CAPTURE \cite{CAPTURE} and CAPability~\cite{liu2025capability}, often evaluate texts by segmenting them into isolated phrases; this fragmentation disrupts the inherent semantic coherence required for chart comprehension.}

 \textcolor{black}{Chart-specific evaluation metrics also face limitations. For instance, ChaTS-Critic~\cite{ChaTS-Critic} evaluated sentence-level faithfulness by leveraging LLMs alongside extracted underlying chart data. However, extracting underlying data directly from chart images, particularly for complex charts, inevitably leads to cascading errors and compromises the reliability of the evaluation.} \textcolor{black}{ChartCap~\cite{lim2025chartcap} evaluates a description by reconstructing a chart from it and comparing the reconstruction with the original. However, this approach requires fine-grained details that are rarely present in real-world descriptions, which typically summarize high-level semantics rather than specify chart details. To address these challenges, we introduce an evaluation framework that systematically assesses the faithfulness and insightfulness of chart descriptions.}

% 现有的图像描述评估指标（如 BLEU、CIDEr 和 METEOR）主要基于候选文本与参考文本之间的 n-gram 片段匹配计算分数。尽管这些指标应用广泛，但其过于依赖表层词汇对齐，难以准确衡量图表描述的事实正确性、信息覆盖度和洞察深度。这种方法可能导致图表描述中存在事实错误或逻辑断层，仍因词汇重叠而获得高分；同时，该类指标对写作风格高度敏感，使得风格各异但内容准确的描述面临不公正的惩罚，无法真实反映模型在图表描述生成任务中的表现。为弥补这一缺陷，CLIPScore 利用 CLIP 模型计算图像与文本的语义相似度，但其主要关注高层语义对齐，难以处理长篇幅且细节丰富的描述。此外，面向通用图像描述的评估方法难以直接迁移至图表描述任务。例如，CAPTURE将描述切分为相对独立的词或短语作为评价单元，而图表描述内部存在较强的语义关联，简单切分会破坏整体语义结构，无法准确反映模型对图表的理解程度及其信息传达质量。在图表描述评估中，ChartCap提出图表描述评估指标Visual Consistency Score (VCS)，通过对比依据描述重构图表与原图之间的相似度来评估图表描述质量。然而，该指标对图表描述粒度要求较高，实际场景中的图表描述多为宏观趋势的语义概括，缺乏支撑图表逆向重构的细粒度信息，适用范围受限。针对这一问题，本研究从准确性、完整性与相关性三个维度构建了多维评估体系，旨在对图表描述质量进行更为全面和系统的评估。

\section{High-quality Chart Description Characterization}

% In this section, we derive the requirements for constructing a high-quality chart description dataset. We summarize the characteristics of high-quality chart descriptions to guide the construction of a comprehensive benchmark dataset.

% \subsection{Requirements}

% \textcolor{black}{Empirical studies show that readers strongly prefer chart descriptions that convey deeper insights quickly grasp the core content of the data and guide them in uncovering deeper insights.~\cite{fourlevel,vistext}. Yet existing benchmark fall systematically short of this standard in two respects.}

\label{sec:3.2}
 \textcolor{black}{Effective chart descriptions enable readers to rapidly grasp key data insights and uncover underlying trends. However, descriptions within current datasets~\cite{lim2025chartcap,wang2025chartinsighter} frequently fail to meet these expectations.} They tend to present visual features as unfocused enumerations. This lack of narrative prioritization forces readers to manually distill critical information, thereby undermining the text's communicative efficacy and imposing an unnecessary cognitive burden~\cite{wiseman2017challenges}. Furthermore, these descriptions rarely integrate domain-specific context, \textcolor{black}{creating a clear mismatch with user preferences, which contrasts with prior findings that both sighted and visually impaired readers prefer descriptions enriched with contextual and semantic information}\cite{fourlevel,vistext}.

 % To address these limitations, we characterize high-quality chart descriptions along two core dimensions: \textbf{Faithfulness} (factual alignment with the data) and \textbf{Insightfulness} (\textcolor{black}{meaningful analytical insights). These dimensions guide the design of our benchmark and evaluation frameworks.} We further decompose Insightfulness via three progressive sub-dimensions: Salient Feature Emphasis for targeted focal selection, Domain-Informed Guidance for interpretive depth, and Chart-Text Complementarity.
 \textcolor{black}{To address these limitations, we characterize high-quality chart descriptions along two core dimensions: \textit{Faithfulness}, which ensures factual alignment with the underlying data, and \textit{Insightfulness}, which requires descriptions to provide meaningful analytical insights. These dimensions guide the design of our benchmark and evaluation frameworks, ensuring that chart descriptions effectively support human analysis. Specifically, \textit{Insightfulness} is further decomposed into three sub-dimensions: Salient Feature Emphasis, Domain-Informed Guidance and Chart–Text Complementarity.}
% \textbf{Salient Feature Emphasis} determines the focal selection of descriptions; \textbf{Domain-Informed Guidance} provides interpretive depth; and \textbf{Chart–Text Complementarity} balances information between the chart and its description.

% we systematically summarize the core characteristics of high-quality chart descriptions(Faithfulness and Insightfulness), insight-driven descriptions and enables a more comprehensive evaluation of computational models, ensuring that the generated description aligns closely with human analytical needs.

% Based on the requirements identified above, we synthesize insights from prior work and propose four quality dimensions that collectively characterize high-quality chart descriptions. These dimensions form a progressive structure from foundational constraint to expressive optimization: \textbf{Factual Accuracy} serves as a necessary prerequisite; \textbf{Salient Feature Emphasis} determines the focal selection of descriptions; \textbf{Domain-Informed Guidance} provides interpretive depth; and \textbf{Chart–Text Complementarity} governs the information division between visual and verbal channels.
% Factual Faithfulness.
\textbf{D1: Factual Accuracy.} \textcolor{black}{A high-quality chart description must faithfully reflect the underlying data in the chart\cite{liu2023autotitle}, as erroneous chart descriptions may mislead users, resulting in flawed decision-making\cite{huang2023zero,chan2023interpretable,kim2024can,huang2024pixels}. Prior work has confirmed that factual hallucinations remain a persistent challenge in automatically generated chart descriptions~\cite{vistext,wang2025chartinsighter,chart-to-text}, underscoring factual accuracy as a fundamental criterion for high-quality descriptions.}

\textbf{D2: Salient Feature Emphasis.} \textcolor{black}{A chart description should prioritize visually salient features, such as extreme values or anomalies, rather than indiscriminately enumerating all visible information. Prior research has shown that when textual emphasis aligns with visually salient features in a chart, readers are more likely to perceive those features as the chart’s central message. Conversely, when a description emphasizes visually inconspicuous and analytically unimportant information without providing additional explanation, readers tend to discount such content and continue to regard the visually salient features as the primary takeaway~\cite{kim2021towards}. However, visual salience should not be treated as the sole criterion, as subtle local variations may be analytically important due to their relevance to a research hypothesis or an underlying mechanism and may therefore warrant explanation.}

\textbf{D3: Domain-Informed Guidance.} \textcolor{black}{A high-quality description should draw on relevant domain knowledge to explain the underlying causes and implications of the observed insights\cite{fourlevel}. Moreover,} a chart may contain multiple insights, yet not all extractable insights carry equal value. A
high-quality chart description should incorporate relevant context and
domain expertise to prioritize the most significant phenomena, articulating why they merit attention and what underlying mechanisms or theoretical implications
they may reflect, rather than enumerating all possible interpretations
indiscriminately~\cite{battle2023we}. Through such domain-grounded and purposeful exposition, the description guides readers from merely observing the data to understanding why the data matter.

% High-quality chart descriptions are characterized by their ability to bridge the semantic gap between observable visual patterns and deeper domain insights~\cite{beyonddescription}. Rather than indiscriminately enumerating all discernible trends, a grounded description demonstrates strict selectivity, prioritizing phenomena that hold theoretical significance~\cite{stokes2022striking}. \textcolor{black}{Furthermore, it offers mechanistic explanations by articulating why specific trends merit attention and how they relate to the broader research context~\cite{battle2023we}. Ultimately, domain grounding elevates the description from a superficial observational log to a domain-specific analytical exposition.}

\textbf{D4: Chart–Text Complementarity.}
 \textcolor{black}{Chart-text complementarity relies on a semantic division of labor: charts excel at presenting numerical relationships and spatial distributions, whereas descriptions should provide high-level synthesis, interpretation, and contextualization. A description fails this complementary role when it mechanically reproduces low-level visual details—such as the narration of axis labels or color encodings~\cite{wang2025characterizing}.} Such overlap degrades synergy into redundancy; it forces readers into constant context-switching between modalities to verify correspondences, inducing a split-attention effect~\cite{latif2021kori,ayres2005split,sweller2005implications} that increases cognitive load without information gain. \textcolor{black}{However, references to visual elements are encouraged when they support higher-level insights, rather than merely restate what is already visible in the chart.}

\begin{figure*}
\includegraphics[width=1\textwidth]
{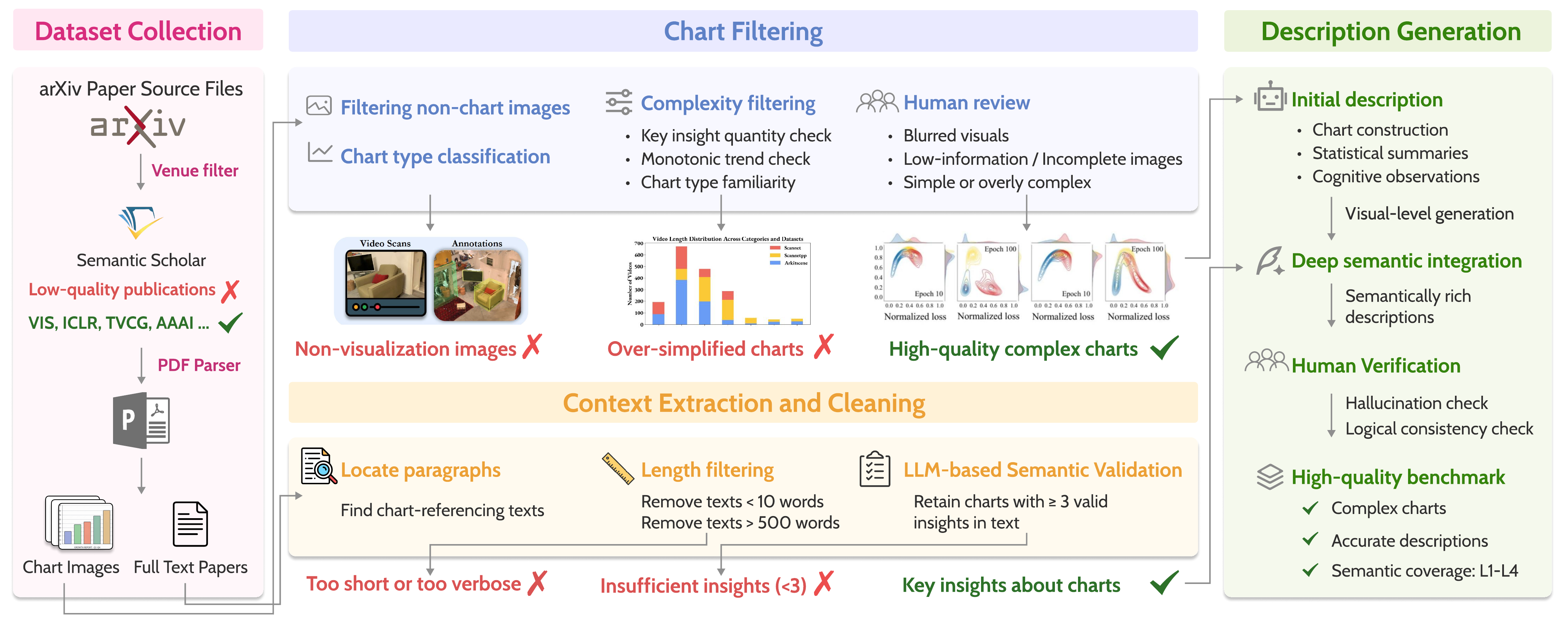}
% \vspace{-5px}
\caption{Overview of the benchmark construction pipeline, consisting of four stages: dataset collection from academic papers, chart filtering and quality control, context extraction and cleaning, and description generation with multi-level semantic coverage.}
 \vspace{-5px}
\label{fig:Dataset Construction}     
\end{figure*}

\section{ChartFI-Bench: A benchmark for chart description}

Based on the characteristics of high-quality chart descriptions (Sec.~\ref{sec:3.2}), we construct a high-quality benchmark containing 896 pairs of charts and corresponding descriptions \textcolor{black}{(Fig.~\ref{fig:Dataset Construction}).}

\subsection{Dataset Construction}

Based on our preliminary study (Sec.\ref{sec:3.2}) and related benchmark works\cite{wang2025chartinsighter,vistext,chart-to-text,chen2024viseval}, we identify three core requirements that the benchmark dataset should meet: 1) \textcolor{black}{\textbf{Diversity in Visual Encodings:}} The dataset should reflect the complexity of real-world visualizations, incorporating a wide taxonomy of chart types, diverse visual marks, and multi-chart visualizations. 2) \textbf{High-Level Analytical Insight:} \textcolor{black}{The description should use domain knowledge to interpret the chart and highlight its most important findings, rather than merely enumerate what it shows. 3) \textbf{Accurate ground truth:} Every claim in a description should be faithfully derived from the underlying data of the chart.}
% \textcolor{black}{To ensure rigorous evaluation of chart description generation, every claim within the ground truth must be faithfully derived from the underlying data of the chart, strictly eliminating external hallucinations unsupported by the chart itself.}

\textbf{Data source and selection.} \textcolor{black}{The arXiv~\cite{clement2019use} repository hosts a massive volume of scientific preprints, offering a raw data pool of authentic, complex visualizations accompanied by detailed, domain-specific explanations.} Consequently, we downloaded all arXiv preprints published from January 2024 to February 2026. \textcolor{black}{To ensure a high standard of data quality, we designed a rigorous filtering mechanism.} We employed Semantic Scholar\cite{kinney2023semantic} to retrieve metadata and publication records, retaining only papers formally accepted at leading venues (e.g., ICLR, AAAI, TVCG). This selection criterion ensured that the charts and their corresponding descriptions had undergone a rigorous peer-review process, thereby guaranteeing a high standard of data quality. This process resulted in a collection of 54,335 papers. We subsequently employed MinerU\cite{niu2026mineru2}  to parse the filtered PDF documents \textcolor{black}{and extract figures together with their surrounding textual context, yielding a total of 302,977 raw figures.}

\textbf{Chart filtering.} \textcolor{black}{The raw figure set contained numerous non-visualization images (e.g., flowcharts) and visually similar charts. To ensure that the retained charts were both visually complex and structurally diverse}, we designed a three-stage filtering pipeline. \textcolor{black}{First, we employed an MLLM to filter the raw images, retaining only valid data visualizations that matched our predefined chart types. Second, we implemented a complexity filter to exclude simplistic charts\cite{ellemose2025eye}.} Specifically, the MLLM assessed whether each candidate contained multiple independent key insights. \textcolor{black}{Charts exhibiting simple data distributions or highly uniform patterns (e.g., a multi-series line chart with an overall upward trend) were classified as simple and removed. Third, three authors conducted a final manual review to remove visually ambiguous, overly simple, or visually similar charts, including those lacking axis labels, legends, or sufficient semantic context.} This step ensured that every retained chart could effectively support the generation of semantically rich descriptions, resulting in a final collection of 1,530 charts from the original 53,660 LLM-extracted candidate charts.

\textbf{Context extraction and cleaning.} Following chart selection, we extracted contextual paragraphs associated with each target chart from the source literature. \textcolor{black}{Specifically, we identified the corresponding chart captions and paragraphs in the main text that referenced the target charts, retaining only textual contexts containing 10--500 words; longer passages were excluded because they often contained substantial information unrelated to the target chart, making them unsuitable for generating focused chart descriptions.} Furthermore, we employed the LLM to conduct a semantic assessment of the candidate texts, retaining only those charts supported by at least three valid data insights. \textcolor{black}{Finally, we extracted chart-relevant insights to provide domain-specific context for subsequent description generation.}

\textbf{Description generation.} \textcolor{black}{First, we prompted GPT-5.2 \cite{gpt} with the chart image to generate a preliminary description covering semantic levels L1--L3~\cite{fourlevel}.} Although this content faithfully reflected the information directly extracted from the chart, \textcolor{black}{it remained limited to visual-level observations and lacked contextual and domain-specific insights (L4). To address this, we provided the previously distilled context in the second stage to guide Gemini-3.0-Pro \cite{gemini} in further refining the initial description. \textcolor{black}{Specifically, the model used the distilled contextual information to identify and highlight the most salient data insights, explain the underlying causes of the observed patterns, and avoid unfocused lists of broad observations.} This effectively directed the narrative toward the conclusions most valuable to the reader.} Ultimately, this process yielded a total of 992 chart descriptions that aligned with the high-quality characteristics defined in Sec.~\ref{sec:3.2} (D2, D3, and D4).

\textbf{Human verification.} Given the risk of hallucination in MLLMs, we implemented a rigorous manual verification process\cite{11264329}. Because the second stage of generation incorporated contextual text from the source literature, the resulting descriptions frequently contained experimental settings or background details not directly inferable from the chart images alone. Therefore, three authors conducted an item-by-item review guided by two core principles: factual fidelity and narrative coherence. To ensure factual fidelity, the authors cross-checked all claims against \textcolor{black}{both the corresponding chart images and the source text}, correcting errors such as hallucinated numerical values and misidentified extrema. \textcolor{black}{Furthermore, we refined the descriptions by removing unverifiable claims and restructuring incoherent content, thereby improving the coherence and reliability of the description.} In total, the manual stages of chart filtering and description verification amounted to approximately 182 person-hours. Ultimately, we constructed a dataset comprising 896 high-quality chart–description pairs.

\begin{figure}
    \centering

    \includegraphics[width=\columnwidth]{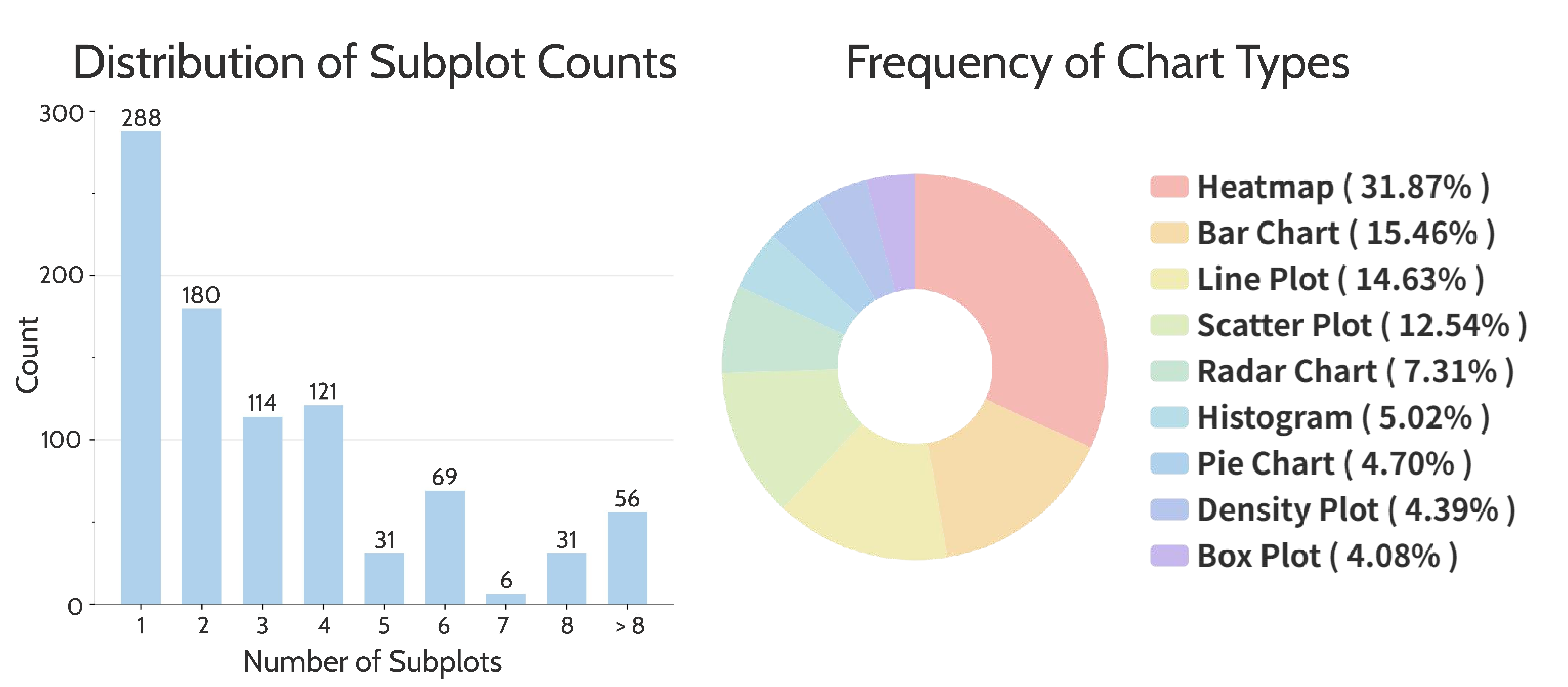}
    \caption{\textcolor{black}{Distribution of subplots per chart (left) and the nine most frequent chart types (right). The complete distribution of chart types is provided in the supplementary material.}}
   \label{fig:chart_type}
   \vspace{-8px}
\end{figure}

\subsection{Dataset Analysis}

\textbf{Visual Diversity.}
ChartFI-Bench covers 13 major chart types, with less frequent types grouped into Others category, providing the broadest coverage among existing benchmarks (Tab.~\ref{tab:dataset_comparison}). \textcolor{black}{Although line and bar charts are the most prevalent chart types in the scientific literature, they tend to be relatively simple in structure. In contrast, heatmaps, which frequently encode high-dimensional and complex data, constitute the largest proportion of our dataset (Fig.~\ref{fig:chart_type}, right), followed by line charts and bar charts.}
Beyond chart type diversity, the benchmark comprises 288 single charts and 608 multi-charts, such as scatter and pie charts in a single plot (Fig.~\ref{fig:chart_type}, left).
The source papers cover diverse domains, predominantly Computer Science (714 charts), followed by Electrical Engineering and Systems Science (85), and Statistics (46). The complete domain breakdown is provided in the supplementary materials.

\textbf{Linguistic Diversity.}
\textcolor{black}{ChartFI-Bench has the longest average description length, at 241 words per description, compared with prior benchmarks (Tab.\ref{tab:dataset_comparison}).} The descriptions span four semantic levels (L1--L4)~\cite{fourlevel}, distinguishing ChartFI-Bench from prior benchmarks that are limited to L1--L3. Tab.~\ref{tab:data fact} presents the distribution of insight types across 12 chart types, highlighting the semantic richness of our descriptions. The complete distribution across all chart types is provided in the supplementary materials.

% \textbf{Human Evaluation.}
% chart construction belongs to L1; value, distribution, extrema, outliers, proportion, and range belong to L2; comparison, trend, correlation, rank, and hierarchy belong to L3; and domain-specific belongs to L4.

\begin{table}[htbp]
\centering
\caption{Distribution of description insights across the 12 chart types.}
\label{tab:data fact}

% 1. 提升字号（比 tiny 大，比 normal 小，适合宽表）
\fontsize{8}{8}\selectfont

% 2. 这里的 tabcolsep 是基础间距，配合 tabular* 的 fill 功能使用
\setlength{\tabcolsep}{0.55pt} 
\resizebox{\columnwidth}{!}{
% 3. 设置行高，保持视觉上的呼吸感（微调至 1.2，因为下方增加了行间距）
\renewcommand{\arraystretch}{1.15} 

% 使用 \columnwidth 确保表格填满当前栏/页宽
% 只要把这里的 \small 改成 \scriptsize，就能单独缩小数字的字号
\begin{tabular*}{\columnwidth}{
l
@{\extracolsep{0.6pt}}
*{12}{>{\centering\arraybackslash\fontsize{6}{8}\selectfont}m{0.45cm}}
@{}
}
    \toprule

    \textbf{Fact Type} & 
    \multicolumn{1}{c}{\raisebox{-1.1ex}{\rotatebox[origin=l]{90}{\textbf{Bar}}}} & 
    \multicolumn{1}{c}{\raisebox{-1ex}{\rotatebox[origin=l]{90}
    {\textbf{Box plot}}}} & 
    \multicolumn{1}{c}{\raisebox{-1ex}{\rotatebox[origin=l]{90}{\textbf{Density}}}} & 
    \multicolumn{1}{c}{\raisebox{-1ex}{\rotatebox[origin=l]{90}{\textbf{Heatmap}}}} & 
    \multicolumn{1}{c}{\raisebox{-1ex}{\rotatebox[origin=l]{90}{\textbf{Histogram}}}} & 
    \multicolumn{1}{c}{\raisebox{-1.1ex}{\rotatebox[origin=l]{90}{\textbf{Line}}}} & 
    \multicolumn{1}{c}{\raisebox{-1ex}{\rotatebox[origin=l]{90}{\textbf{Pie}}}} & 
    \multicolumn{1}{c}{\raisebox{-1ex}{\rotatebox[origin=l]{90}{\textbf{Radar}}}} & 
    \multicolumn{1}{c}{\raisebox{-1ex}{\rotatebox[origin=l]{90}{\textbf{Scatter}}}} & 
    \multicolumn{1}{c}{\raisebox{-1.1ex}{\rotatebox[origin=l]{90}{\textbf{Bubble}}}} & 
    \multicolumn{1}{c}{\raisebox{-1ex}{\rotatebox[origin=l]{90}{\textbf{Violin}}}} &
    \multicolumn{1}{c}{\raisebox{-1ex}{\rotatebox[origin=l]{90}{\textbf{Area}}}}\\
    \midrule
    \textbf{Chart Construction}\quad\quad 
    & \cellcolor{blue!30}214 & \cellcolor{blue!20}53  & \cellcolor{blue!20}60  & \cellcolor{blue!40}485  & \cellcolor{blue!20}73  & \cellcolor{blue!30}224 & \cellcolor{blue!20}63  & \cellcolor{blue!30}126 & \cellcolor{blue!30}214 & \cellcolor{blue!10}33& \cellcolor{blue!10}27 & \cellcolor{blue!10}27 \\ \addlinespace[0.5pt]
    \textbf{Value}
    & \cellcolor{blue!50}966 & \cellcolor{blue!30}144 & \cellcolor{blue!20}71  & \cellcolor{blue!50}1726 & \cellcolor{blue!20}77  & \cellcolor{blue!40}441 & \cellcolor{blue!30}389 & \cellcolor{blue!40}458 & \cellcolor{blue!30}383 & \cellcolor{blue!30}117 & \cellcolor{blue!20}80 & \cellcolor{blue!10}36\\ \addlinespace[0.5pt]
    \textbf{Distribution }
    & \cellcolor{blue!30}226 & \cellcolor{blue!20}80  & \cellcolor{blue!30}290 & \cellcolor{blue!40}570  & \cellcolor{blue!30}210 & \cellcolor{blue!30}135 & \cellcolor{blue!20}71  & \cellcolor{blue!20}89  & \cellcolor{blue!30}367 & \cellcolor{blue!10}32 & \cellcolor{blue!20}64 & \cellcolor{blue!10}17 \\ \addlinespace[0.5pt]
   \textbf{Extrema } 
    & \cellcolor{blue!30}190 & \cellcolor{blue!10}31   & \cellcolor{blue!10}21   & \cellcolor{blue!30}270  & \cellcolor{blue!10}40  & \cellcolor{blue!30}137 & \cellcolor{blue!10}26   & \cellcolor{blue!20}91  & \cellcolor{blue!20}84 & \cellcolor{blue!10}27 & \cellcolor{blue!10}31 & \cellcolor{blue!10}23 \\ \addlinespace[0.5pt]
   \textbf{Range    }
    & \cellcolor{blue!30}130 & \cellcolor{blue!20}67  & \cellcolor{blue!10}22   & \cellcolor{blue!30}206  & \cellcolor{blue!10}25   & \cellcolor{blue!20}83  & \cellcolor{blue!10}15   & \cellcolor{blue!10}30   & \cellcolor{blue!30}126 & \cellcolor{blue!10}22 & \cellcolor{blue!10}43 & \cellcolor{blue!10}1 \\ \addlinespace[0.5pt]
     \textbf{Outlier} 
     & \cellcolor{blue!10}13   & \cellcolor{blue!10}8    & \cellcolor{blue!10}6    & \cellcolor{blue!20}51   & \cellcolor{blue!10}16   & \cellcolor{blue!10}9    & \cellcolor{blue!10}3    & \cellcolor{blue!10}4    & \cellcolor{blue!10}25  & \cellcolor{blue!10}4 & \cellcolor{blue!10}4  & \cellcolor{blue!0}0 \\ \addlinespace[0.5pt]
     \textbf{Proportion}  
     & \cellcolor{blue!20}71  & \cellcolor{blue!10}6    & \cellcolor{blue!10}4    & \cellcolor{blue!10}33 & \cellcolor{blue!10}7    & \cellcolor{blue!10}10   & \cellcolor{blue!10}22   & \cellcolor{blue!10}1  & \cellcolor{blue!10}12 & \cellcolor{blue!10}9 & \cellcolor{blue!10}2  & \cellcolor{blue!10}9\\ \addlinespace[0.5pt]
    \textbf{Comparison}
    & \cellcolor{blue!30}345 & \cellcolor{blue!30}111 & \cellcolor{blue!20}95  & \cellcolor{blue!40}515  & \cellcolor{blue!30}104 & \cellcolor{blue!30}325 & \cellcolor{blue!20}77  & \cellcolor{blue!30}174 & \cellcolor{blue!30}235 & \cellcolor{blue!10}49 & \cellcolor{blue!10}38 & \cellcolor{blue!10}24\\ \addlinespace[0.5pt]
    \textbf{Trend}        
    & \cellcolor{blue!30}307 & \cellcolor{blue!20}89  & \cellcolor{blue!10}47  & \cellcolor{blue!40}511  & \cellcolor{blue!20}64  & \cellcolor{blue!40}606 & \cellcolor{blue!10}35   & \cellcolor{blue!20}59  & \cellcolor{blue!30}222 & \cellcolor{blue!10}23 & \cellcolor{blue!10}28 & \cellcolor{blue!30}124 \\ \addlinespace[0.5pt]
    \textbf{Correlation}
    & \cellcolor{blue!20}61  & \cellcolor{blue!10}25   & \cellcolor{blue!10}16   & \cellcolor{blue!30}334  & \cellcolor{blue!10}15   & \cellcolor{blue!20}69  & \cellcolor{blue!10}11   & \cellcolor{blue!10}6    & \cellcolor{blue!30}126 & \cellcolor{blue!10}9 & \cellcolor{blue!10}4  & \cellcolor{blue!10}27\\ \addlinespace[0.5pt]
    \textbf{Rank}     
    & \cellcolor{blue!30}167 & \cellcolor{blue!10}38   & \cellcolor{blue!10}16   & \cellcolor{blue!30}218  & \cellcolor{blue!10}27   & \cellcolor{blue!20}94  & \cellcolor{blue!30}206 & \cellcolor{blue!30}123 & \cellcolor{blue!30}131 & \cellcolor{blue!10}4 & \cellcolor{blue!10}11 & \cellcolor{blue!10}11 \\ \addlinespace[0.5pt]
    \textbf{Hierarchy}   
    & \cellcolor{blue!10}7    & \cellcolor{blue!0}0    & \cellcolor{blue!10}7    & \cellcolor{blue!10}34    & \cellcolor{blue!0}0    & \cellcolor{blue!10}4    & \cellcolor{blue!20}100 & \cellcolor{blue!0}0    & \cellcolor{blue!10}3    & \cellcolor{blue!0}0 & \cellcolor{blue!0}0  & \cellcolor{blue!10}4\\ \addlinespace[0.5pt]
    \textbf{Domain-Specific}
    & \cellcolor{blue!30}422 & \cellcolor{blue!30}120 & \cellcolor{blue!30}123 & \cellcolor{blue!50}852  & \cellcolor{blue!30}131 & \cellcolor{blue!30}374 & \cellcolor{blue!30}120 & \cellcolor{blue!30}188 & \cellcolor{blue!30}333 & \cellcolor{blue!20}52 & \cellcolor{blue!10}47 & \cellcolor{blue!10}42 \\
    \bottomrule
\end{tabular*}
}
\vspace{-3pt} 
\end{table}

\section{Evaluation Metrics}

In this section, we introduce our evaluation framework comprising four metrics: Faithfulness, Coverage, Informativeness, and Acuity.
\label{sec:metric}

\subsection{Data Fact Formulation}

\label{sec:datafact}

Chart descriptions convey diverse insight types, such as trends, extrema, and comparisons. \textcolor{black}{To support fine-grained analysis while abstracting away linguistic variations, we first decompose each description into fine-grained chart insights. Content at L2 and L3 is represented as structured data facts, whereas original sentences at L1 and L4 are decomposed into fine-grained sentences, each conveying a single insight.} Adapted from DataShot\cite{wang2019datashot}, we formalize each data fact as a 6-tuple: $(\mathit{type},\ \mathit{parameters},\ \mathit{measure}(s),\ \mathit{context},\ \mathit{breakdown}(s),\ \mathit{focus})$, where $type$ specifies the category of the insight (e.g., trend, rank or proportion) based on existing taxonomies\cite{wang2019datashot,shao2025narrative,das2025charts,shen2022visual}; $parameters$ describe the specific characteristics of the fact; $measure(s)$ identify the fields treated as dependent variables; $context$ defines the data subspace; $breakdown(s)$ specifies the grouping dimensions; and $focus$ identifies the highlighted groups within those dimensions. 

\textcolor{black}{To enable reliable automatic matching of semantically equivalent data facts despite variations in natural language phrasing, we adopt predefined extraction rules for each $\mathit{type}$.} Specifically, trend facts are restricted to directional states (e.g., \textsc{Increase}, \textsc{Decrease}, \textsc{Peak}); comparison and correlation facts encode relational directions (e.g., \textsc{Positive}, \textsc{Negative}) alongside the compared entities; and distribution facts capture structured degree-state pairs (e.g., \textsc{Highly Clustered}). \textcolor{black}{This standardized representation allows semantically equivalent insights to be consistently matched despite differences in their surface forms (as detailed in Sec.~\ref{sec:coverage}). These structured data facts support the subsequent fine-grained evaluation of chart descriptions.}

\subsection{Faithfulness}
\label{sec:faithfulness}
 % \textcolor{black}{To assess the fidelity of chart descriptions with respect to their underlying data(D1), we systematically design four candidate verification methods and conduct a controlled comparative study to identify the most effective approach. }

 \textcolor{black}{To validate \textbf{D1} (Factual Accuracy), we evaluate the Correctness of chart descriptions to their underlying data. We design four candidate verification methods and conduct a comparative study to identify the most effective approach. These four methods arise from a $2 \times 2$ experimental design crossing two factors: input representation (original descriptions vs.\ \textcolor{black}{fine-grained chart insights}) and verification mechanism (\textit{MLLM-as-a-Judge} vs.\ code-based verification)}. Specifically, the \textit{MLLM-as-a-Judge} mechanism employs the MLLM to directly assess the chart image against either the original description (\textit{Method~1}) or \textcolor{black}{fine-grained chart insights} extracted from that description (\textit{Method~2}). In contrast, the code-based approach consists of two steps: the MLLM first extracts the underlying data from the chart image, and the LLM subsequently generates executable code to verify either the original description (\textit{Method~3}) or each \textcolor{black}{chart insight} (\textit{Method~4}) using the data.

We curated a test set of 16 chart descriptions spanning 13 chart types (e.g., radar chart, heatmap and bubble chart). Sampled from the outputs of various proprietary and open-source models (e.g., GPT-5.4\cite{gpt}, Qwen3.5-27B\cite{qwen3.5} and InternVL3.5-14B\cite{wang2025internvl3}), these descriptions yield a total of 938 distinct verification points. The ground truth labels were annotated by two authors based on the chart images. \textcolor{black}{We evaluate performance using precision, recall, and F$_1$-score. As shown in Fig.~\ref{fig:correctness_performance}, \textit{Method 1} achieves the best overall performance with an F$_1$-score of 0.91. Regarding the verification mechanism, direct MLLM judgment (\textit{Methods 1 and 2}) consistently surpasses code-based verification (\textit{Methods 3 and 4}). Furthermore, using the original descriptions (\textit{Methods 1 and 3}) yields higher precision than using \textcolor{black}{fine-grained chart insights} (\textit{Methods 2 and 4}).}

To obtain deeper insights into the failure modes associated with each method, we perform a case-by-case error attribution analysis on all error cases, categorizing these errors into five types (Fig. \ref{fig:correctness_performance}) : 1) \textbf{Verification Omission}, where the evaluated method overlooks critical error information present in the description; 2) \textbf{Code Logic Error}, occurring when the generated verification code executes successfully but relies on flawed judgment logic (e.g., \textcolor{black}{verifying a cluster by checking whether at least one point falls within a region, rather than whether a sufficient number of points concentrate there}); 3) \textbf{Extraction Loss}, denoting instances where the extraction process for data facts omits critical details or semantics from the original description \textcolor{black}{(e.g., ``significantly increase'' is reduced to ``increases'', causing the verifier to miss degree-level errors)}; 4) \textbf{MLLM Misjudgment}, wherein the model's comprehension or reasoning about the chart image is fundamentally erroneous; 5) \textbf{Data Extraction Error}, wherein the model extracts an erroneous data table from the chart, consequently causing code verification to yield incorrect results.

The distribution of these failure modes varies significantly across the four evaluated methods. 
\textcolor{black}{Notably, Code Logic Error occurs only in \textit{Methods 3} and \textit{4}; nevertheless, it still accounts for 20.3\% of the total errors.} This suggests that, \textcolor{black}{although code-based verification can improve verification accuracy, LLM-generated verification code often exhibits insufficient logical robustness, introducing errors that undermine this advantage. Likewise, data fact extraction aims to enable finer-grained evaluation by decomposing descriptions into atomic units; however, information loss during extraction can reduce the effectiveness of such fine-grained evaluation.} For example, during data fact extraction, the phrase ``significantly increase'' may be extracted simply as the ``increases''.
When the error of the description lies in misjudging the degree of change rather than the direction, such information loss causes the verification mechanism to overlook the error. This explains why the recall of \textit{Method 4} (0.73) falls below that of \textit{Method 1} (0.88). \textit{Method 1} directly compares the chart image against its corresponding description and achieves the best overall performance; we therefore adopt it for faithfulness verification in our evaluation framework. The \textbf{Faithfulness} score is defined as follows:
\begin{equation}
\mathrm{Faithfulness}
 = 1 - \frac{N_{\text{error}}}{N_{\text{total}}}
\end{equation}
where $N_{\text{total}}$ and $N_{\text{error}}$ denote the total number of \textcolor{black}{chart insights} in the description and the number of erroneous \textcolor{black}{chart insights}, respectively.

\begin{figure}
    \centering
    \includegraphics[width=0.9\columnwidth]{figs/Faithfulness.pdf}
    \caption{Performance comparison of four methods for faithfulness evaluation (top) and their corresponding error analysis (bottom).}
    \label{fig:correctness_performance}
    \vspace{-1px}
\end{figure}

\subsection{Coverage}
\label{sec:coverage}
\textcolor{black}{Coverage assesses whether a model-generated description captures the key insights by comparing it against the reference description.} We decompose both the reference and generated descriptions into fine-grained \textcolor{black}{chart insights}. Formally, let $R={r_1, r_2, \dots, r_n}$ denote the set of \textcolor{black}{chart insights} extracted from the reference description and $M={m_1, m_2, \dots, m_k}$ the set extracted from the model-generated description. \textcolor{black}{For each reference fact $r_i$, we compute the matching score $\phi(r, m)$ against every generated \textcolor{black}{insight} $m_j$. A reference–generated \textcolor{black}{insight} pair is considered eligible for matching if its matching score exceeds a predefined threshold $\tau$. Accordingly, the set of eligible \textcolor{black}{insight} pairs is defined as:}

\begin{equation}
\mathcal{E}
=
\left\{
(r_i, m_j) \in R \times M
\;\middle|\;
\phi(r_i, m_j) \geq \tau
\right\}.
\label{eq:eligible_pairs}
\end{equation}

% \begin{equation}
% k =
% \left|
% \left\{
% r_i \in R
% \mid
% \exists\, m_j \in M,\ 
% \phi(r,m) \geq \tau
% \right\}
% \right|.
% \label{eq:coverage}
% \end{equation}

\textcolor{black}{For L1 and L4 insights, we directly compute the semantic similarity between the sentences using Qwen3-Embedding-0.6B\cite{qwen3.5}. For L2 and L3 insights represented as structured data facts,} the matching score $\phi(r,m)$ is computed by comparing the corresponding fields of the reference data fact $r$ and the generated data fact $m$, and aggregating the resulting matching scores across fields using field-specific weights:

\begin{equation}
\phi(r, m) = \sum_{d \in D} w_d \cdot \sigma_d(r, m)
\end{equation}
where $D$ denotes data fact field set defined in Sec. \ref{sec:datafact}, $w_d$ is the weight assigned to field $d$, and $\sigma_d(r,m)$ measures the matching score between $r$ and $m$ on the corresponding field $d$. We use field-specific matching criteria to compute scores as follows:

\textbf{Type-Specific Fact Matching.} \textcolor{black}{We apply type-specific matching rules according to the data fact type.} For numeric types (e.g., value, rank, and proportion), we evaluate approximate numerical equality within a configurable tolerance. For trend facts, we extract directional keyword sequences from the relevant \textit{parameters} and assess agreement through longest common subsequence (LCS) matching. For comparison and correlation facts, we verify \textcolor{black}{the consistency of the specified relation (e.g., \textsc{Positive} versus \textsc{Negative}) while accounting for the ordering of the involved \textcolor{black}{measures}. For comparison facts, reversing the \textcolor{black}{measures} order requires the relation to be inverted accordingly.} For distribution facts, \textcolor{black}{we separately compare the degree and state parameters (e.g., \textsc{Highly} and \textsc{Clustered}) and assign partial scores when only one of the two parameters matches.} The complete matching specification is provided in the supplemental material.

\textbf{Schema-grounded fact normalization.} \textcolor{black}{Several fields in a data fact, such as \textit{breakdown(s)} and \textit{focus}, contain variable names or expressions derived from the chart description.} Consequently, the same underlying variable may be expressed differently across descriptions (e.g., ``mean average precision'' versus ``mAP'', or ``accuracy'' versus ``acc score''), rendering direct comparison between field values unreliable. To mitigate such lexical variation, we extract a schema of canonical variable names from each chart (e.g., axis labels and data categories) and normalize the values of these fields to their corresponding canonical entries in the schema before matching. For field values that cannot be mapped to the schema, we fall back to calculating semantic similarity scores using Qwen3-Embedding-0.6B\cite{qwen3.5}.

\textbf{Cross-Type Semantic Matching.} \textcolor{black}{The same underlying chart insight may be represented by different data fact types.} For example, a maximum value can appear as an extremum fact (``A achieves the highest score'') or a rank fact (``A ranks first''). \textcolor{black}{Restricting alignment to facts of the same type would therefore miss such semantically equivalent cases.} To address this issue, we define a set of cross-type equivalence groups (e.g., extremum, rank) and implement dedicated alignment rules with type-specific conversion logic. This enables semantically equivalent facts encoded under different fact types to be matched correctly.

\textcolor{black}{Additionally, we introduce two additional mechanisms to handle special cases during fact matching.} First, we dynamically renormalize the field weights when both the reference fact and the model-generated fact have an N/A value for the same field. Specifically, the weight assigned to this inapplicable field is proportionally redistributed among the remaining applicable fields, preventing it from affecting the overall matching score. Second, rather than matching facts sequentially, we compute the matching score $\phi(r_i, m_j)$ for all reference--model fact pairs, sort the scores in descending order, and greedily assign matches subject to a one-to-one constraint and a minimum threshold $\tau$. \textcolor{black}{This strategy reduces the likelihood that an early, lower-quality match prevents a better subsequent pairing.}

Based on the resulting alignment, we compute three metrics. \textcolor{black}{Let $N$ denote the number of matched \textcolor{black}{insight} pairs, and $|M|$ and $|R|$ the numbers of model-generated and reference \textcolor{black}{chart insights}, respectively}:

\begin{equation}
\text{Precision} = \frac{N}{|M|}, \quad
\text{Recall} = \frac{N}{|R|}, \quad
\text{F}_1 = \frac{2 \cdot \text{Precision} \cdot \text{Recall}}{\text{Precision} + \text{Recall}}
\label{eq:coverage_metrics}
\end{equation}

\textcolor{black}{Recall measures the proportion of reference insights successfully captured by the model-generated description.} Precision complements this by measuring the proportion of generated \textcolor{black}{insights} that can be matched to the reference \textcolor{black}{insights}. The $\text{F}_1$ score summarizes both aspects by balancing Precision and Recall through their harmonic mean. \textcolor{black}{We therefore define \textbf{Coverage} as the $\mathrm{F}_1$ score in Eq.~\ref{eq:coverage_metrics}.} Through extensive empirical calibration, we set the matching threshold at $\tau = 0.7$ and assign the weights of the dimensions as $\{0.3, 0.3, 0.2, 0.1, 0.1\}$ to \textit{parameters}, \textit{measure(s)}, \textit{context}, \textit{breakdown(s)}, and \textit{focus}, respectively.

\subsection{Informativeness}
\label{sec:informativeness}
The complementarity principle (D4) established in Sec.~\ref{sec:3.2} states that high-quality chart descriptions should not reproduce low-level visual details that are readily apparent from the chart, but should instead focus on higher-level insights. To operationalize this qualitative principle into a quantifiable evaluation criterion, we propose the Informativeness Metric. The core approach decomposes a generated chart description into \textcolor{black}{fine-grained chart insights} across different semantic levels~\cite{fourlevel}, categorized from L1 to L4. We assign a base weight $b_l$ to each level. Through extensive experiments across diverse types of charts and styles of description, we empirically determine these weights to be $b_l \in \{1, 6, 7, 7\}$.

However, \textcolor{black}{fixed weighting schemes do not account for the structural variations among chart descriptions with different levels of complexity, potentially leading to biased evaluations of description quality}. Descriptions of complex charts typically allocate a larger proportion of content to the visual encoding and layout details of L1, thereby reducing the relative proportion of higher-level insights. Conversely, descriptions of structurally simple charts require less space for the exposition of visual elements, which allows the analytical content of L3 and L4 to occupy a greater proportion. To address this limitation, we introduce a mechanism for context-adaptive weight modulation driven by the distribution of levels in the reference description. This mechanism enables the metric to automatically calibrate the relative importance of each level according to the analytical emphasis of the specific chart. Let $p_l$ denote the proportion of units at level $l$ in the reference description. The context-aware weight is defined as:
\vspace{-0.1cm}
\begin{equation}
\tilde{w}_l = b_l  \exp(p_l)
\end{equation}

\textcolor{black}{When a specific insight level accounts for a larger proportion of the reference description ($p_l$ is large), the exponential term increases the corresponding weight non-linearly.} Conversely, when a level is absent from the reference ($p_l = 0$), the exponential term equals one, and the weight reverts to the base weight $b_l$. \textcolor{black}{This design avoids a potential limitation of directly assigning weights from the reference distribution: reference descriptions may fail to cover all insight levels due to annotation constraints, such as limited description length or annotators' varying emphasis.} The resulting raw weights are normalized:
\vspace{-0.1cm}
\begin{equation}
w_l = \frac{b_l \exp(p_l)}{\sum_{k=1}^{L} b_k \exp(p_k)}
\end{equation}
Given a sequence of semantic units $Y = \{y_1, y_2, \dots, y_{|Y|}\}$ extracted from the candidate description, \textcolor{black}{where $y_i$ denotes the $i$-th \textcolor{black}{chart insight} and $|Y|$ denotes the total number of \textcolor{black}{chart insights}}, we define \textbf{Informativeness} as the average weight across all \textcolor{black}{chart insights}:
\vspace{-0.15cm}
\begin{equation}
\mathrm{Informativeness} = \frac{1}{|Y|} \sum_{i=1}^{|Y|} w_l(y_i)
\end{equation}

\begin{table*}[t]
\centering
% 这里的 9.5pt 是字号，11pt 是行距（一般行距比字号大 1.2-1.5 倍）
% 如果您的正文字号是 11pt，这里可以改成 \fontsize{10.5pt}{12pt}\selectfont
\fontsize{9pt}{9pt}\selectfont 
\caption{\textcolor{black}{Performance comparison across models on reference-based metrics and the four proposed metrics.}}
\vspace{-0.05cm}
\label{tab:model_results}
\setlength{\tabcolsep}{5pt}
\renewcommand{\arraystretch}{1.15}

\begin{tabular*}{\textwidth}{@{\extracolsep{\fill}}lcccccccc}
\toprule
\multirow{2}{*}{\textbf{Model}} & \multicolumn{4}{c}{\textbf{Reference-based Metrics}} & \multicolumn{4}{c}{\textbf{ChartFI}} \\
\cmidrule(lr){2-5} \cmidrule(lr){6-9}
 & \textbf{BLEU} & \textbf{METEOR} & \textbf{ROUGE} & \textbf{BLEURT} 
 & \textbf{Faithfulness} & \textbf{Coverage} & \textbf{Informativeness} & \textbf{Acuity} \\
\midrule
GPT-5.4          & \textbf{0.0731} & 0.2894          & \textbf{0.2206} & -0.4832          & \textbf{0.9501} & \textbf{0.3358} & \textbf{0.3188} & \textbf{3.1233} \\

Gemini-3-Flash   & 0.0723          & \textbf{0.3076} & 0.2163          & \textbf{-0.4342} & 0.9471          & 0.3145          & 0.2762          & 3.0719 \\

Claude-Sonnet-4.6   & 0.0651          & 0.2852 & 0.2139          & -0.4870 & 0.8712          & 0.3038          & 0.3033          & 2.8919 \\

Qwen3.5-plus     & 0.0615          & 0.2821          & 0.1939          & -0.4803          & 0.8638          & 0.3059          & 0.2854          & 2.9930 \\

Qwen3.5-27B      & 0.0574          & 0.2687          & 0.1938          & -0.4849          & 0.8591          & 0.2901          & 0.2747          & 2.8249 \\

Llama-4-Scout-17B-16E & 0.0450          & 0.2541          & 0.1915          & -0.4745          & 0.7348          & 0.2539          & 0.3058          & 2.1833 \\

InternVL3.5-14B  & 0.0538          & 0.2578          & 0.2001          & -0.4764          & 0.8134          & 0.2461          & 0.2667          & 1.7589 \\

\bottomrule
\end{tabular*}
\vspace{-5pt} %
\end{table*}

\subsection{Acuity}
\label{sec:acuity}
We propose Acuity to measure whether a chart description can bridge the gap between \emph{what the chart shows} and \emph{what the chart means}. \textcolor{black}{We operationalize this metric through five sub-dimensions, each evaluated by Gemini-3.1-Pro\cite{gemini} on a 5-point Likert scale.} The five sub-dimensions are defined as follows:

\textbf{Domain Knowledge Accuracy \& Relevance.}
This sub-dimension assesses whether the description incorporates domain concepts that are technically accurate and directly pertinent to the provided chart. It evaluates the effective mapping of visual elements to established phenomena, problems, or theoretical constructs within the field, rather than relying on generic or loosely associated terminology\cite{lu2026domaincqa}.

\textbf{Knowledge Integration Efficiency.}
Possessing relevant knowledge alone is insufficient; it must be utilized efficiently. 
This sub-dimension evaluates whether the introduced domain context demonstrably enhances the reader's understanding of specific chart phenomena. This avoids both under-explanation (where necessary knowledge is absent) and over-elaboration (where accurate but dispensable background obscures the core analytical insights).

\textbf{Insight Prioritization.}
A chart typically presents multiple valid insights, yet these insights vary significantly in value for research or decision-making\cite{kim2021towards}. Identifying the findings that deserve prominence requires domain expertise, such as recognizing that an unexpected drop in performance at a specific scale is more consequential than a general upward trend. This sub-dimension assesses whether the description leverages domain knowledge to emphasize the most analytically significant findings and articulate why they merit attention, rather than merely reporting visually prominent features without selection informed by domain knowledge.

\textbf{Etiological Explanation.}
Beyond identifying \emph{what} patterns exist, understanding \emph{why} they emerge represents the most critical application of domain knowledge\cite{hoque2025domain}. This sub-dimension evaluates whether the description provides plausible, domain-grounded explanations for the observed phenomena, and whether it delineates a clear boundary between interpretations supported by evidence and speculative assertions.

\textbf{Multivariate Synthesis.} \textcolor{black}{Charts with multiple variables or subplots often require joint interpretation, as important relationships may only become apparent through comparisons across variables or subplots~\cite{wang2024charxiv}.} Crucially, determining the variables or conditions that warrant comparison, and understanding the significance of their joint patterns, depends fundamentally on domain expertise. This sub-dimension evaluates whether the description synthesizes information across variables or subplots to generate higher-order insights informed by domain knowledge, rather than treating them as isolated observations.

\textcolor{black}{The overall \textbf{Acuity} score aggregates the evaluations across these sub-dimensions. Specifically, \textbf{Acuity} is defined as the mean of the valid sub-dimension scores:}
\vspace{-0.2cm}
\begin{equation}
    \mathrm{Acuity} = \frac{1}{N} \sum_{i=1}^{N} A_i
    \label{eq:acuity}
\end{equation}
\noindent\textcolor{black}{where $N$ denotes the number of applicable sub-dimensions out of the total five. If the chart does not involve multiple variable dimensions, the Multivariate Synthesis sub-dimension is excluded from the aggregation, reducing the denominator to $N = 4$. Otherwise, $N = 5$.}

\section{Evaluation}

We evaluate \textcolor{black}{seven MLLMs, validate our metrics, and compare metric changes using author-written chart descriptions as references.}

\subsection{Setup}

\label{sec: evaluation models}
\textbf{Models.} To comprehensively assess the capabilities of chart summarization across diverse model architectures, we evaluate both proprietary and open-source MLLMs. Regarding proprietary models, we benchmark GPT-5.4\cite{gpt}, Gemini-3-Flash\cite{gemini}, \textcolor{black}{Claude-Sonnet-4.6\cite{claude}}, and Qwen3.5-plus\cite{qwen3.5}. As for open-source models, we select Qwen3.5-27B\cite{qwen3.5}, \textcolor{black}{Llama-4-Scout-17B-16E\cite{adcock2026llama}}, and InternVL3.5-14B\cite{wang2025internvl3}. Following established practices in prior work~\cite{wang2024charxiv,chen2024viseval}, we set the temperature to 0 and the top-$p$ value to 1 for all models to ensure deterministic and reproducible outputs.

% Following previous setups, we set the temperature  0 top=1 to achieve optimal results.

\textbf{Metrics.} We employ two categories of evaluation metrics: traditional natural language generation (NLG) metrics and our four proposed metrics (Sec.~\ref{sec:metric}). Regarding the traditional NLG metrics, we adopt BLEU\cite{papineni2002bleu} for n-gram precision, METEOR\cite{banerjee2005meteor} for alignment quality with the awareness of synonyms and stemming, ROUGE\cite{lin2004rouge} for measuring recall-oriented lexical overlap, and BLEURT\cite{sellam2020bleurt} (specifically BLEURT-base-128) for the learning-based assessment of fluency and semantic adequacy. For the proposed metrics, we evaluate Faithfulness and Acuity using Gemini-3.1-Pro\cite{gemini} as the adjudicator model.

\textbf{Prompt.} We utilize the prompt ``Write a description of the chart(s) in a paragraph of at least 150 words'' for all evaluated models. Extensive experiments indicate that without such a length constraint, the models tend to generate overly brief descriptions that fail to capture the complete informational content of the charts. By specifying a requirement for length, we encourage the models to produce more comprehensive descriptions that better reflect the true capabilities of the models in chart understanding. This strategy is consistent with the prompting protocol adopted by CAPTURE\cite{CAPTURE}.

\textbf{Qualitative analysis.} To better understand the model limitations reflected by Acuity, we examined the generated analytical descriptions of representative proprietary and open-source models, including GPT-5.4\cite{gpt}, Qwen3.5-plus\cite{qwen3.5}, and Qwen3.5-27B\cite{qwen3.5}, under extended thinking mode. \textcolor{black}{From the charts for which all three models achieved Acuity scores below 3, we sampled 30 charts across different levels of average Acuity score, while ensuring diversity in domains and visualization types. We then manually analyzed the generated descriptions for these charts to identify recurring limitations.} Detailed case analyses are provided in the supplementary material.

\subsection{Main Results}
\label{sec:automatic evaluation}
\textcolor{black}{\textbf{MLLMs often produce hallucinations when generating chart descriptions.} As shown in Tab.~\ref{tab:model_results}, the top-performing model achieves a Faithfulness score of 0.9501. In contrast, open-source models exhibit a noticeable performance drop, with Llama-4-Scout-17B-16E scoring 0.7348, highlighting a substantial gap between the leading proprietary and open-source models. These results suggest that hallucinations remain a persistent issue.} For example, when a chart encodes multiple variables, MLLMs often confuse the mappings between colors and categories. Moreover, MLLMs often misread specific numerical values from chart elements, such as the heights of bars or the positions of lines, especially in the absence of explicit data labels. These low-level reading errors can lead to higher-level analytical failures, affecting the correctness of subsequent trend characterization, comparison reasoning, and extrema identification. \textcolor{black}{Beyond such visual reading errors, hallucinations also emerge in domain-grounded interpretation. As shown in Tab.~\ref{tab:domain_dimensions}, MLLMs achieve only moderate scores on the Domain Knowledge Accuracy \& Relevance dimension; specifically, Gemini-3-Flash scores 3.18, while GPT-5.4 scores 3.10. This suggests that MLLMs may also introduce inaccurate or contextually irrelevant domain explanations when interpreting chart insights.}

\begin{table}[t!]
\centering
\fontsize{8pt}{8pt}\selectfont 
\caption{\textcolor{black}{Model performance across the five sub-dimensions of Acuity. Acc., Integ., Insight, Etiol., and Multi. refer to Domain Knowledge Accuracy \& Relevance, Knowledge Integration Efficiency, Insight Prioritization, Etiological Explanation, and Multivariate Synthesis, respectively.}}
% \vspace{-0.1cm}
\label{tab:domain_dimensions}
\renewcommand{\arraystretch}{1.1}
\begin{tabular*}{0.96\columnwidth}{@{\extracolsep{\fill}}lccccc@{}}
\toprule
\textbf{Model} & \textbf{Acc.} & \textbf{Integ.} & \textbf{Insight} & \textbf{Etiol.} & \textbf{Multi.} \\
\midrule
GPT-5.4          & 3.10 & \textbf{3.31} & \textbf{3.32} & 2.44 & \textbf{3.45} \\
Gemini-3-Flash   & \textbf{3.18} & 3.17 & 3.20 & \textbf{2.54} & 3.27 \\
Claude-Sonnet-4.6   & 2.99 & 2.96 & 2.98 & 2.47 & 3.06 \\

Qwen3.5-plus     & 3.08 & 3.18 & 3.03 & 2.38 & 3.31 \\
Qwen3.5-27B      & 2.86 & 2.99 & 2.76 & 2.38 & 3.13 \\
Llama-4-Scout & 2.18 & 2.23 & 2.16 & 2.01 & 2.34 \\
InternVL3.5-14B & 1.85 & 1.91 & 1.73 & 1.32 & 1.99 \\
\bottomrule
\end{tabular*}
\vspace{-5pt} %
\end{table}

\textcolor{black}{\textbf{MLLMs struggle to prioritize the most analytically meaningful insights.} Across all evaluated MLLMs, Coverage and Informativeness scores remain relatively low, indicating that MLLMs do not merely omit key information from charts but also struggle to determine which chart details are analytically worth emphasizing. In practice, MLLMs show a strong preference for generating L1 content. Although such descriptions may be factually accurate, they remain superficial and contribute little to explaining the main findings conveyed by the chart. This tendency is particularly pronounced in open-source models. Additionally, rather than prioritizing analytically meaningful insights, they often attend to visually salient features, such as the largest radar-chart regions or the darkest heatmap cells, and treat them as important without verifying whether they correspond to the central analytical findings. However, visual salience does not necessarily imply analytical importance; determining whether a visually prominent feature is analytically meaningful often requires an understanding of the variables, the chart context, and relevant domain knowledge. Current MLLMs, however, often fail to incorporate such knowledge into their reasoning process. This limitation is reflected in the relatively modest scores on Acuity’s Insight Prioritization dimension, ranging from 1.73 to 3.32 (Tab.~\ref{tab:domain_dimensions}).  }

\textcolor{black}{More importantly, MLLMs often fail to synthesize multiple dimensions into a coherent analytical account. They can often identify that trends differ across subplots, but they rarely explain these differences by considering all relevant dimensions jointly. Instead, they focus on one or two dimensions while overlooking additional ones that jointly determine the observed insights. As a result, the generated descriptions capture only partial insights rather than complete explanations of the underlying multivariate relationships. This limitation remains evident even for the strongest proprietary models, with GPT-5.4 achieving a Multivariate Synthesis score of 3.45 (Tab.~\ref{tab:domain_dimensions}).}

\textcolor{black}{\textbf{MLLMs struggle to incorporate domain knowledge into chart interpretation.} Although MLLMs can often identify salient visual patterns, they frequently struggle to connect these patterns with relevant domain knowledge. On the Etiological Explanation dimension (Tab.~\ref{tab:domain_dimensions}), even the best-performing model, Gemini-3-Flash, achieves only 2.54, suggesting that MLLMs rarely explain what the observed observations imply or why they matter in the corresponding domain. More importantly, even when MLLMs introduce relevant domain context, they often fail to translate it into concrete explanations that clarify specific chart findings and help readers better understand the chart. For example, MLLMs may state that ``larger models benefit from stronger reasoning ability,'' but they often do not connect this claim to the specific performance in the chart. This limitation is also reflected in the Knowledge Integration Efficiency scores reported in Tab.~\ref{tab:domain_dimensions}, which range from only 1.91 to 3.31. As a result, their descriptions may remain factually plausible, but they are often analytically hollow, offering generic commentary rather than domain-informed interpretive insights.}

\begin{figure}
    \centering
    \includegraphics[width=\columnwidth, height=0.5\textheight, keepaspectratio]{figs/Automated-Authored.pdf}
    \caption{\textcolor{black}{Comparison of MLLM performance using pipeline-generated references (Automated) and author-written references (Authored) across evaluation metrics (a)-(g); frequency distributions of the four semantic levels in Automated and Authored references (h).}}
    \vspace{-8pt}
    \label{fig:autherd_references}
\end{figure}

\subsection{Metric Analysis}

\textbf{\textcolor{black}{Consistency with human evaluation.}} To validate the effectiveness of the proposed automatic evaluation metrics, we conducted a human evaluation study. We randomly sampled \textcolor{black}{80} chart descriptions from our benchmark, generated by seven models (described in Sec.~\ref{sec: evaluation models}). The domain distribution of the sampled descriptions was guided by the full benchmark, resulting in a CS to non-CS ratio of 54:26. Six domain experts (aged 20–30) with experience in reading and producing chart descriptions participated in the study. Each description was assigned to an expert with matching domain expertise. Before the evaluation process, we thoroughly explained the scoring criteria of the four evaluation dimensions to the experts. The experts first reviewed representative examples to resolve ambiguities and ensure a consistent understanding of the rating standards. Then, each expert independently scored the descriptions on a 5-point Likert scale across all four dimensions of the evaluation metrics. To quantify the consistency between the proposed automatic metrics and the human judgments, we adopted the Spearman rank correlation coefficient (SRCC), a widely acknowledged metric utilized in prior studies~\cite{chen2024viseval,recall2016beyond,fu2019visualization}. SRCC measures the degree to which the predicted ranking of the scores aligns with the human-assigned ranking, with values falling within the range of [-1, 1]: a value of 1 indicates perfect rank agreement, whereas -1 indicates a completely inverted ordering.

As shown in Tabs.~\ref{tab:srcc_results} and~\ref{tab:acuity_gpt_gemini_human}, the SRCC values across all four dimensions exceed \textcolor{black}{0.69}, demonstrating a strong positive correlation between the automatic metrics and human assessments. Furthermore, we assessed inter-expert consistency using four experts, including two from computer science and two from electrical energy storage systems. For each domain, the two corresponding experts independently scored the same six samples, and their pairwise SRCC values were averaged across domains, yielding average inter-expert correlations above 0.8 (Tab.~\ref{tab:srcc_results}). Specifically, a detailed breakdown of the Acuity metric (Tab.~\ref{tab:acuity_gpt_gemini_human}) reveals that all of its five sub-dimensions also \textcolor{black}{align closely with human scores. Together, these results support the validity of the proposed metrics for automatic evaluation of chart descriptions.} \textcolor{black}{Additionally, to mitigate single-judge bias and verify metric robustness, we further evaluated the same set of 80 chart descriptions using GPT-5.4\cite{gpt}. As shown in Tab.~\ref{tab:acuity_gpt_gemini_human}, the pairwise SRCC among GPT-5.4, Gemini-3.1-Pro\cite{gemini}, and humans across all five Acuity sub-dimensions exceeds 0.69, confirming the metric's robustness to the choice of judge model.}

\textcolor{black}{\textbf{Consistency across domains.}
Given that the benchmark is predominantly composed of arXiv-sourced CS charts, we partition our evaluation results into CS and non-CS subsets to examine potential domain bias and its impact on model evaluation (Fig.~\ref{fig:breakdown_of_cs_noncs}). The evaluated metrics exhibit broadly consistent trends across the two subsets, with no obvious domain bias.}

% \textcolor{black}{\textbf{Comparison with other metrics.}}

\begin{figure}
    \centering
    \includegraphics[width=\columnwidth]{figs/cs-nocs.pdf}
    \caption{\textcolor{black}{Comparison of MLLM performance across evaluation metrics between CS ($n=714$) and non-CS ($n=182$) domains.}}
    \label{fig:breakdown_of_cs_noncs}
    % \vspace{-12pt}
\end{figure}

% To further investigate whether the reference descriptions in our benchmark accurately reflect the original authors' intent, we conducted an elicitation study.

\subsection{\textcolor{black}{Evaluation with Author-written References}}
\label{sec:Evaluation with Expert-authored References}
\textcolor{black}{Our benchmark construction pipeline uses chart-related text written by the original authors to construct chart descriptions. However, the resulting descriptions may still not fully recover the authors’ communicative intent. This limitation may affect the benchmark’s ability to evaluate how well MLLMs generate chart descriptions aligned with the authors’ intended insights. To examine this potential limitation, we contacted the authors of 441 charts from our benchmark, successfully collecting 49 author-written descriptions. Upon manual inspection, we excluded 6 responses that were not valid chart descriptions (e.g., those including contextual details uninferable from the chart). The remaining 43 descriptions were utilized as ground-truth references, alongside the synthesized descriptions generated by our pipeline, to re-evaluate the MLLMs. Since the Faithfulness metric is evaluated reference-free, our subsequent analysis focuses on the remaining seven reference-based metrics (detailed in Sec.~\ref{sec: evaluation models}).}

\begin{table}[t]
\centering
\caption{SRCC across four evaluation dimensions for agreement between the proposed metrics and human evaluation, and inter-expert agreement.}
% \vspace{-0.05cm}
\label{tab:srcc_results}
\resizebox{\linewidth}{!}{
\renewcommand{\arraystretch}{1.1}
\begin{tabular}{lcccc}
\toprule
\textbf{Evaluators} & \textbf{Faithfulness} & \textbf{Coverage} & \textbf{Informativeness} & \textbf{Acuity} \\
\midrule
Our method vs. Human    & 0.809 & 0.785 & 0.786 & 0.858 \\
Expert vs. Expert & 0.857 & 0.800 & 0.800 & 0.829 \\
\bottomrule
\end{tabular}
}
% \vspace{-10pt} %
\end{table}

\begin{table}[t!]
\centering
\caption{\textcolor{black}{Pairwise SRCC among human and model evaluations across Acuity and its five sub-dimensions. GPT and Gemini denote evaluations by GPT-5.4\cite{gpt} and Gemini-3.1-Pro\cite{gemini}, respectively.}}
% \vspace{-0.2cm}
\label{tab:acuity_gpt_gemini_human}
\renewcommand{\arraystretch}{1.1}
% 下面这行是核心：将内容缩放至当前列宽 (\columnwidth)
\resizebox{\columnwidth}{!}{% 
\begin{tabular}{lcccccc}
\toprule
\textbf{Evaluators} & \textbf{Acc.} & \textbf{Integ.} & \textbf{Insight} & \textbf{Etiol.} & \textbf{Multi.} & \textbf{Acuity}\\
\midrule
Gemini vs. Human   & 0.790 & 0.842 & 0.826 & 0.790 & 0.767 & 0.858\\
GPT vs. Human     & 0.734 & 0.709 & 0.764 & 0.704 & 0.693 & 0.784\\
GPT vs. Gemini     & 0.898 & 0.864 & 0.857 & 0.792 & 0.819 & 0.908\\
\bottomrule
\end{tabular}
} % 这里是 \resizebox 的结束括号
\vspace{-5pt}
\end{table}

\textcolor{black}{
As shown in Fig.~\ref{fig:autherd_references}-h, the semantic level distribution (L1--L4) of author-written references differs from pipeline-generated references. Authors typically include a higher proportion of L1 content, and emphasize the advantages of their proposed methods while devoting less attention to other data dimensions. This compositional shift alters the dynamic weight allocation across L1--L4 within the Informativeness metric, leading to corresponding score variations (Fig.~\ref{fig:autherd_references}-b). However, the relative performance rankings of the evaluated MLLMs remain stable. Notably, overall Coverage scores decline when evaluated against author-written references (Fig.~\ref{fig:autherd_references}-a). This reveals a critical evaluation bias: our pipeline-generated references are denser in L2 and L3 content, which models like GPT-5.4 and Claude-Sonnet-4.6 excel at capturing, yielding higher Coverage scores on the pipeline-generated references. Conversely, models like Qwen3.5-27B and Qwen3.5-plus, despite their lower Informativeness scores, achieve high Coverage against author references by successfully matching a substantial volume of L1 content. Regarding the Acuity metric (Fig.~\ref{fig:autherd_references}-c), the relative rankings of the MLLMs remain stable despite minor cross-model fluctuations. Furthermore, under traditional reference-based metrics, the scores of all evaluated models drop significantly. This is because chart authors employ highly contextualized and linguistically diverse phrasing that exposes the limitations of standard n-gram matching metrics, rendering them inadequate for assessing the capabilities in generating chart descriptions.}

\section{discussion}
In this section, we discuss the limitations and broader implications of our work, along with potential directions for future research.

\textbf{\textcolor{black}{Tailoring Descriptions to Author Intent and Reader Backgrounds.}} \textcolor{black}{Our description generation framework builds on the four-level semantic content hierarchy\cite{fourlevel} and uses chart-related text written by the authors as reference. However, as discussed in Sec.~\ref{sec:Evaluation with Expert-authored References}, automatically generated chart descriptions may still exhibit subtle deviations from expert-authored descriptions in their semantic-level composition and analytical emphasis. This suggests that a single reference description may not fully capture the authors’ intent or the expectations of readers with different levels of background knowledge.} Future work could explore intent-aware and audience-aware chart description generation, where the depth and emphasis of each semantic level are adjusted according to the intended communicative goal and target audience.

% Our description framework extends the four-level semantic content hierarchy\cite{fourlevel} by incorporating Level 4 content, which introduces contextual and domain-specific insights. However, readers with varying levels of background knowledge may hold different expectations for chart descriptions\cite{hsu2023gpt}. Future work could explore the generation of adaptive descriptions that tailor the depth and emphasis of each semantic level based on the profile of the intended reader. This objective could be achieved through user studies that elicit preferences across different levels of expertise, or through controllable generation techniques that enable users to specify the desired level of detail.

\textcolor{black}{ 
\textbf{Advancing chart description generation with ChartFI.}
Our evaluation framework can provide long-term support for assessing semantic quality in chart descriptions, particularly hallucination and insightfulness. Our current quantitative results and failure analysis reveal key limitations of current models, offering insights for designing targeted improvement strategies such as chain-of-thought prompting, multi-agent workflows, and evaluation-guided optimization. In addition, our dataset can serve as a seed set for automatically deriving synthetic training data to fine-tune future models.
}

\textbf{Limitations of Evaluation Metrics.} Our framework provides a systematic evaluation of chart descriptions across multiple quality dimensions. However, as discussed in Sec.~\ref{sec:faithfulness}, detecting fine-grained hallucinations remains challenging. Future work could fine-tune verification models with explicit error taxonomies to improve fine-grained hallucination detection. Moreover, Coverage metric relies on LLMs to extract data facts from references and the generated descriptions. However, semantically similar insights may be expressed differently and assigned to different types, leading to potential mismatches during the alignment phase. Future research could investigate more robust semantic matching strategies to improve the reliability of the coverage evaluation. Finally, the Informativeness metric may be sensitive to the number of extracted data facts, as linear normalization can disproportionately penalize descriptions containing more data facts. Future work could explore sub-linear normalization to better balance informativeness and description length.

\textbf{Dataset Scale and Diversity.} Our dataset is primarily composed of scientific charts sourced from arXiv. Given that our filtering criteria prioritize chart complexity, and computer science dominates the literature on arXiv while charts from other domains are often comparatively simple, the dataset predominantly consists of visually complex charts from the computer science domain. \textcolor{black}{Future work could apply our construction pipeline to broader sources and domains, while involving domain experts to contribute and annotate high-quality descriptions, thereby expanding both the scale and diversity of the dataset\cite{8017645,9552226,9904480,10292610}.}

\section{conclusion}
We introduce ChartFI-Bench, a comprehensive benchmark designed for the evaluation of chart descriptions along two dimensions: Faithfulness and Insightfulness. We first systematically summarize the characteristics of high-quality chart descriptions. Based on these principles, we construct a rigorously curated dataset of 896 chart--description pairs sourced from arXiv. We further propose a four-dimensional evaluation framework to assess the Faithfulness and Insightfulness of the descriptions. Experiments on mainstream MLLMs demonstrate the effectiveness of the proposed framework and expose shared weaknesses across current models. Our framework and evaluation results can advance research in high-quality chart description generation.

\section*{ACKNOWLEDGMENTS}

The authors want to thank the reviewers for their suggestions. This work was supported by the National Natural Science Foundation of China (No. 62472099), the Fundamental and Interdisciplinary Disciplines Breakthrough Plan of the Ministry of Education of China (No. JYB2025XDXM904), the AI for Science Program of the Shanghai Municipal Commission of Economy and Informatization (Grant No. 2025-GZL-RGZN-BTBX-02028), and the Ji Hua Laboratory S\&T Program (No. X250881UG250).

\bibliographystyle{abbrv-doi-hyperref}

\bibliography{ref}

% \appendix % You can use the `hideappendix` class option to skip everything after \appendix
% \crefalias{section}{appendix} % this is to make sure that cleverref switches to referring to Appx. X from here on

% \section{About Appendices}
% Refer to \cref{sec:appendices_inst} for instructions regarding appendices.

% \section{Troubleshooting}
% \label{appendix:troubleshooting}

% \subsection{ifpdf error}

% If you receive compilation errors along the lines of \texttt{Package ifpdf Error: Name clash, \textbackslash ifpdf is already defined} then please add a new line \verb|\let\ifpdf\relax| right after the \verb|\documentclass[journal]{vgtc}| call.
% Note that your error is due to packages you use that define \verb|\ifpdf| which is obsolete (the result is that \verb|\ifpdf| is defined twice); these packages should be changed to use \verb|ifpdf| package instead.

% \subsection{\texttt{pdfendlink} error}

% Occasionally (for some \LaTeX\ distributions) this hyper-linked bib\TeX\ style may lead to \textbf{compilation errors} (\texttt{pdfendlink ended up in different nesting level ...}) if a reference entry is broken across two pages (due to a bug in \verb|hyperref|).
% In this case, make sure you have the latest version of the \verb|hyperref| package (i.e.\ update your \LaTeX\ installation/packages) or, alternatively, revert back to \verb|\bibliographystyle{abbrv-doi}| (at the expense of removing hyperlinks from the bibliography) and try \verb|\bibliographystyle{abbrv-doi-hyperref}| again after some more editing.

\end{document}